%% file: main.tex
\documentclass{article}

\usepackage[final]{nips_2016}

\usepackage[utf8]{inputenc} % allow utf-8 input
\usepackage[T1]{fontenc}    % use 8-bit T1 fonts
\usepackage{hyperref}       % hyperlinks
\usepackage{url}            % simple URL typesetting
\usepackage{booktabs}       % professional-quality tables
\usepackage{amsfonts}       % blackboard math symbols
\usepackage{nicefrac}       % compact symbols for 1/2, etc.
\usepackage{microtype}      % microtypography

\usepackage{comment}
\usepackage{graphicx}
\usepackage{subfig}
\usepackage{wrapfig,lipsum,booktabs}
\usepackage{multirow}
\usepackage{hhline}

\usepackage{todonotes}

\usepackage{xspace}

\usepackage{mathrsfs}

\usepackage{amsthm}

\usepackage{subfig}% http://ctan.org/pkg/subfig
\usepackage{booktabs}% http://ctan.org/pkg/booktabs

\usepackage{bm} 

\usepackage{wrapfig}

\usepackage{epstopdf}
\AppendGraphicsExtensions{.tif}

\newcommand{\Prv}{$\mathcal{P}$\xspace}
\newcommand{\Vrf}{$\mathcal{V}$\xspace}
\newcommand{\sys}{SafetyNets\xspace}

\newtheorem{theorem}{Theorem}[section]

\newtheorem{lemma}[theorem]{Lemma}

\title{SafetyNets: Verifiable Execution of Deep Neural Networks on an Untrusted Cloud}

\author{
  Zahra Ghodsi, Tianyu Gu, Siddharth Garg \\
  New York University\\
  \texttt{\{zg451, tg1553, sg175\}@nyu.edu} \\
}

\begin{document}

\maketitle

\input{abstract}
\input{introduction}

\input{background}

\input{safetynets}
\input{eval}

\bibliography{main}
\bibliographystyle{abbrv}

\input{proofs}

%\newpage
%\appendix
%\include{proofs}
\end{document}

%% file: abstract.tex
\begin{abstract}
  Inference using deep neural networks is often outsourced to the cloud since 
  it is a computationally demanding task. However, this raises a fundamental issue of trust. 
  How can a client be sure that the cloud has performed inference correctly? A lazy cloud provider  
  might use a simpler but less accurate model to reduce its own computational load, or worse, 
  maliciously modify the inference results sent to the client. We propose \sys, a framework that 
  enables an untrusted server (the cloud) to provide a client with a short mathematical proof of 
  the correctness of inference tasks that they perform on behalf of the client. Specifically, \sys 
  develops and implements a specialized \emph{interactive proof} (IP) protocol for verifiable   
  execution of a class of deep neural networks, i.e., those that can be represented as arithmetic circuits. 
  Our empirical results on three- and four-layer deep neural networks 
  demonstrate the run-time costs of 
  \sys for both the client and server are low. 
  %is 
  %low-cost,  
  \sys detects any incorrect computations of the neural network by the untrusted server 
  with high probability, 
  while achieving state-of-the-art accuracy on the MNIST digit recognition ($99.4\%$) and 
  TIMIT speech recognition tasks ($75.22\%$).
  %: the server's overhead in generating proofs and the client's work in 
  %verifying proofs is low.
\end{abstract}

%% file: introduction.tex
\section{Introduction} \label{intro}

Recent advances in deep learning have shown 
that 
multi-layer neural networks can achieve
state-of-the-art performance on a wide range of
machine learning tasks. 
However, training and performing inference 
on neural networks can be computationally expensive.  
%and are often outsourced to the cloud. 
For this reason, 
several commercial vendors have begun offering 
``machine learning as a service" (MLaaS) solutions that allow clients 
to \emph{outsource} machine learning computations, both training and 
inference, to the cloud. 

%Outsourcing computations to a 
%third party, in this case the cloud, 
%comes at the expense of trust, i.e.,  
While promising, the MLaaS model (and outsourced computing, in general) 
raises immediate security concerns, specifically relating to  
the 
\emph{integrity} (or correctness) of computations performed 
by the cloud and the \emph{privacy} of the client's data~\cite{papernot2016towards}. 
This paper focuses on the former, i.e., the question of 
integrity.  Specifically, how can a client perform
%outsource the 
%task of 
inference using a deep neural network on an untrusted
cloud, while obtaining strong assurance that the cloud has performed   
inference correctly? 

%Cite 
%https://arxiv.org/pdf/1611.03814.pdf

Indeed, there are compelling reasons for 
a client to be wary of a third-party cloud's computations. 
For one, 
the cloud has a financial incentive to be ``lazy." 
A lazy cloud might use a simpler but less accurate
model, for instance, a single-layer instead of a multi-layer 
neural network, to reduce its computational costs. Further 
the cloud could be compromised by malware that 
modifies the results sent back to the 
client with malicious intent. For instance, the cloud 
might always mis-classify a certain digit 
in a digit recognition task, or allow unauthorized access 
to certain users 
in a face recognition based authentication system.

%Trust issues with cloud computing
%http://www.netsec.ethz.ch/publications/papers/gennaro_gentry_parno_verifiable.pdf

%Survey paper
%https://link.springer.com/chapter/10.1007/978-3-319-53798-6_8

The security risks posed by cloud computing have spurred 
theoretical  
advances in the area of 
\emph{verifiable computing} (VC)~\cite{walfish2015verifying}. 
The idea 
is to enable a client to 
\emph{provably} (and \emph{cheaply}) 
verify that an untrusted 
server has performed computations correctly. 
To do so, the 
server provides to the client (in addition to the result of computation) 
a \emph{mathematical} {proof} of the correctness of the result. 
The client rejects, with high probability,  
any incorrectly computed results
(or proofs) provided 
by the server, while always accepting 
correct results (and corresponding proofs)
\footnote{Note that the \sys is not intended to and cannot catch any inherent 
mis-classifications 
due to the model itself, only those that result from incorrect computations of the model by the server.}.
VC techniques aim for the following desirable 
properties: the size of the 
proof should be small, 
the client's verification effort must be lower than 
performing the computation locally, 
and the server's effort 
in generating proofs should not be too high.

The advantage of proof-based VC  
is that it provides  
\emph{unconditional}, mathematical guarantees on the integrity 
of computation performed by the server.  
Alternative solutions for verifiable execution 
require the client to 
make trust \emph{assumptions} that are hard for the client to independently 
verify. 
Trusted platform modules~\cite{gennaro2010non}, for instance,  
require the client to place trust on the hardware manufacturer, and 
assume that the hardware is 
tamper-proof. Audits 
based on the server's execution time~\cite{monrose1999distributed} 
require
precise knowledge of the server's hardware configuration and assume, 
for instance, that the server is not over-clocked.

%Compared to , can provi
%\emph{unconditional} security guarantees, \todo{insert}
%as opposed to that make restrictive trust assumptions. 

%In this paper, we propose  
%a framework to 
%verify the execution of deep neural networks on 
%untrusted clouds based 

\begin{wrapfigure}{l}{0.6\textwidth}
\centering
    \includegraphics[width=0.55\textwidth]{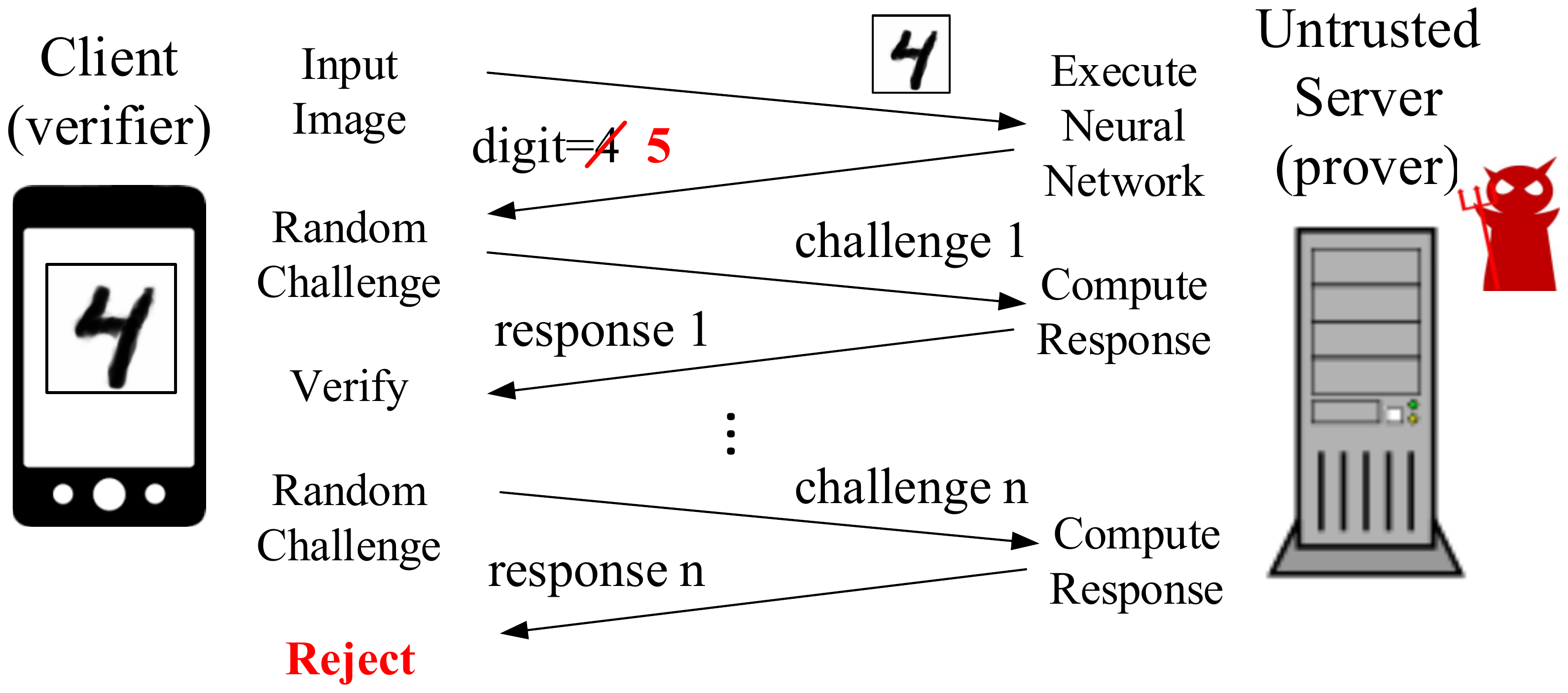}
    \caption{High-level overview of the \sys IP protocol. In this example, an untrusted server intentionally changes the classification output from $4$ to $5$.
    %The client sends the input image to the server which in turn feeds it to the neural network and returns the result. The interactive proof protocol works by several rounds of challenges and responses and the result is only accepted if the server's responses throughout the protocol are consistent.
    }
\label{fig:ip}
\end{wrapfigure}

The work in this paper leverages 
powerful 
VC techniques referred to as 
interactive proof (IP) systems~\cite{cmt,goldwasser2008delegating, thaler2013time,vu2013hybrid}. An IP system consists 
of two entities, a prover (\Prv), i.e., the untrusted 
server, and a verifier (\Vrf), i.e., the client.  The framework is illustrated in 
Figure~\ref{fig:ip}. 
The verifier sends the prover an input $\bm{x}$, say a batch of 
test images, and 
asks the prover
to compute a function $\bm{y} = f(\bm{x})$. 
In our setting, $f(.)$ is a trained 
multi-layer neural network that is known to both the verifier and 
prover, 
and $\bm{y}$ is the neural network's classification output 
for each image in the batch. 
The prover performs the computation and
sends the verifier a purported result $\bm{y'}$ (which is not 
equal to $\bm{y}$ if the prover cheats). 
The verifier and prover then engage in $n$ 
rounds of interaction. In each round, 
the verifier sends the 
prover a randomly 
picked {challenge}, 
and the prover provides a 
response based on the 
IP protocol. The verifier \emph{accepts} 
that  $\bm{y'}$ is indeed equal to $f(\bm{x})$ 
if it is 
satisfied with the prover's response in each round, and \emph{rejects} otherwise.
%(a few numbers from a 
%finite field, as we see shortly), 

%Deployments for general purpose computing too costly. 
%However, certain types of computations are practical. 
%Justin Thaler's work --- distinct, matmult, builds up on 
%IP system. 
%As shown in Fig., 
%verifier (V), prover (P). 
%Over multiple rounds; randomly chosen challenge; response; rejects. 

A major criticism of IP systems (and, indeed, all existing 
VC techniques) when used for verifying general-purpose computations  
is that the prover's overheads are 
\emph{large}, often orders of magnitude 
more than just computing $f(\bm{x})$~\cite{walfish2015verifying}. 
Recently, however, 
Thaler~\cite{thaler2013time} showed that 
certain types of computations
admit IP protocols 
with highly efficient verifiers and provers, which
lays the foundations 
for the 
specialized IP protocols
for deep neural networks 
that we develop in this paper.

%Several theoretical frameworks for verifiable computing 
%have been proposed in literature, and can roughly be categorized as
%based either 
%on interactive proof (IP) systems~\cite{}, or 
%crpytographic arguments~\cite{}\todo{Is FHE based technique a third class or part of crypto args? Check MWs paper}. 

%IP, cryptographic arguments, FHE extension
%unconditional guarantees, 
%unlike TPM require trust the hardware manufacturer and 
%suseeptibly to physical tampering
%provide unconditional, mathematical guarantees.
%However, 

\paragraph{Paper Contributions.} 
This paper introduces \sys, a new (and to the 
best of our knowledge, the first)  
approach
for verifiable execution 
of deep neural networks on untrusted clouds. 
Specifically, \sys composes a new, specialized  
IP protocol for 
the neural network's activation layers 
with Thaler's IP protocol for matrix multiplication 
to achieve end-to-end verifiability, dramatically 
reducing the bandwidth costs versus a naive solution that verifies the 
execution of each layer of the neural network separately. 
%that is, 
%\todo{does not verify each layer separately, saves b/w}

\sys
applies to a certain class of neural networks that 
can be represented as {arithmetic circuits} that  
perform computations over finite fields  
(i.e., integers modulo a large prime $p$). 
Our implementation of 
\sys addresses several practical 
challenges in this context, including the choice of the prime  
$p$, its relationship to accuracy of the 
neural network, and to the  
verifier and prover run-times.  

Empirical evaluations on the MNIST digit recognition and 
TIMIT speech recognition tasks illustrate that \sys 
enables practical, low-cost verifiable outsourcing of deep neural network
execution without compromising classification accuracy. 
Specifically, the 
client's execution time is $8\times$-$80\times$ lower 
than executing the network locally, 
the server's overhead in generating proofs
is less than $5\%$, and the client/server exchange less 
than $8$ KBytes of data during the IP protocol.  
\sys's security guarantees ensure that a client can detect 
any incorrect computations performed by a malicious server  
with probability vanishingly close to $1$. 
At the same time, \sys achieves state-of-the-art 
classification accuracies of $99.4\%$ and $75.22\%$ 
on the MNIST and TIMIT datasets, respectively.

%% file: background.tex
\section{Background}\label{prelims}

%Interactive proofs~\cite{babai1985trading} extend proof systems in complexity theory to be random and interactive. In these probabilistic proof systems, a \emph{prover} (server) tries to convince a \emph{verifier} (client) of the correctness of a mathematical assertion. Figure~\ref{fig:ip} demonstrates an interactive proof system for an image recognition task. The client and server engage in a protocol during which the client will generate random challenges and the server computes and sends back responses. If the server's responses are consistent throughout the protocol, the client accepts the result and otherwise rejects it.
In this section, we begin by reviewing necessary background on IP systems, and then
describe the restricted class of neural networks
(those that can be represented as arithmetic circuits) that \sys handles. 

\subsection{Interactive Proof Systems}\label{sec:ip}
Existing IP systems proposed in literature~\cite{cmt,goldwasser2008delegating, thaler2013time,vu2013hybrid} 
use, at their heart, 
a protocol referred to as the 
%The interactive proof method used in this paper and previous work~\cite{cmt,vu2013hybrid} uses a technique 
%the 
sum-check protocol~\cite{lund1992algebraic} that we describe
here in some detail, and then discuss 
its applicability in verifying 
general-purpose computations 
expressed as arithmetic circuits.

\paragraph{Sum-check Protocol} 
Consider a $d$-degree $n$-variate 
polynomial $g(x_{1},x_{2},\ldots,x_{n})$, 
where each variable
$x_{i} \in \mathbb{F}_{p}$ ($\mathbb{F}_{p}$ is the set of all natural numbers 
between zero and $p-1$, for a given prime $p$) and    
$g: \mathbb{F}_{p}^{n} \rightarrow \mathbb{F}_{p}$. 
The prover \Prv seeks to prove the following claim:
\begin{equation}
   y =  \sum_{x_{1} \in \{0,1\}} \sum_{x_{2} \in \{0,1\}} \ldots \sum_{x_{n} \in \{0,1\}} g(x_{1},x_{2},\ldots,x_{n})
   \label{eq:claim1}
\end{equation}
that is, the sum of $g$ evaluated at $2^n$ points is $y$. 
\Prv and \Vrf now engage in a sum-check protocol 
to verify this claim. In the first round of the protocol, 
\Prv sends the following unidimensional polynomial
\begin{equation}
   h(x_1) =  \sum_{x_{2} \in \{0,1\}} \sum_{x_{3} \in \{0,1\}} \ldots \sum_{x_{n} \in \{0,1\}} g(x_{1},x_{2},\ldots,x_{n})
   \label{eq:hx1}
\end{equation}
to \Vrf in the form of its coefficients. 
\Vrf checks if $h(0)+h(1) = y$. If yes, it proceeds, otherwise it rejects \Prv's claim. 
Next, \Vrf picks a random 
value 
$q_{1} \in \mathbb{F}_{p}$ and evaluates 
$h(q_1)$ which, based on Equation~\ref{eq:hx1}, yields
a new claim:
\begin{equation}
   h(q_1) =  \sum_{x_{2} \in \{0,1\}} \sum_{x_{3} \in \{0,1\}} \ldots \sum_{x_{n} \in \{0,1\}} g(q_{1},x_{2},\ldots,x_{n}).
   \label{eq:claim2}
\end{equation}
%The claim in Equation~\ref{eq:claim2} has the
%same form as the claim in in Equation~\ref{eq:claim1} but with one fewer variable.
%in that 
%it involves the claim about the sum of a $b-1$-variate polynomial 
%over $2_{b-1}$ points (note that variable $x_1$ has been 
%replaced with a constant $q_1$). 
\Vrf now recursively 
calls the sum-check protocol to verify this new claim. 
By the final round of the sum-check protocol, 
%\Vrf 
%has sent $n$ random numbers to \Prv, i.e., $q_1$, $q_2$, 
%ldots $q_n$. In this round, 
\Prv returns the value  
$g(q_{1},q_{2},\ldots,q_{n})$ and the 
\Vrf checks if this value is correct by evaluating the 
polynomial by itself. If so, \Vrf accepts the original 
claim in Equation~\ref{eq:claim1}, otherwise it 
rejects the claim.

\begin{lemma}~\cite{arora2009computational}
\Vrf rejects an incorrect claim by \Prv
with probability 
greater than 
%$\left( 1 - \frac{d}{p} \right) _{n} \approx 
$\left( 1 - \epsilon\right)$ where $\epsilon = \frac{nd}{p}$ 
is referred to as the \bf{soundness error}.% for $p>>d$.
\end{lemma}

\begin{comment}
One set of realistic values with which we 
instantiate the sum-check protocol is 
$d=2$, $b=10$ and 
$p=2_{31}-1$. For these values, 
\Vrf accepts an incorrect claim by \Prv with a probability 
of $\approx \frac{1}{2_{27}}$, an 
infinitesimally small number.

\paragraph{Remarks} Two observations 
are in order. First, the security of the sum-check 
protocol hinges on the fact that \Prv does not know, 
in advance, \Vrf's 
random choices for future rounds. This allows \Vrf to 
(eventually)
catch \Prv in an inconsistency if it lied about a prior 
claim. Second, observe that if \Vrf would need 
to perform exponentially many ($2^b$) evaluations
of $g$ if it chooses to verify the claim in Equation~\ref{eq:claim1} by itself (without the \Prv). 
The sum-check protocol reduces \Vrf's work to a single 
evaluation of $g$, and $O(b)$ evaluations of a single 
variable polynomial. 
\end{comment}

\paragraph{IPs for Verifying Arithmetic Circuits}  
In their seminal work, Goldwasser et al.~\cite{goldwasser2008delegating} 
demonstrated how sum-check can be used to verify the execution of 
arithmetic circuits using an IP protocol now referred to as GKR. 
An arithmetic circuit is a directed acyclic graph of computation over 
elements of a finite field $\mathbb{F}_{p}$
in which each node can perform either \emph{addition} or \emph{multiplication} 
operations (modulo $p$). While we refer the reader to 
~\cite{goldwasser2008delegating} for further details of GKR, one important 
aspect of the protocol bears mention.  

GKR organizes nodes of an arithmetic circuit into \emph{layers}; 
starting with the circuit inputs, the outputs of 
one layer feed the inputs of the next. 
The GKR proof protocol operates backwards from the 
circuit outputs to its inputs. 
Specifically, GKR uses sum-check to reduce 
the prover's assertion about the 
circuit output into an assertion about the inputs of the output 
layer. This assertion is then reduced to an assertion about the inputs of the 
penultimate layer, and so on. 
The protocol continues iteratively till
the verifier is left with an assertion about the 
circuit inputs, which it checks on its own.
The layered nature of GKR's prover aligns 
almost perfectly with the structure of 
a multi-layer neural network 
and motivates the 
use of an IP system 
based on GKR for \sys.

\subsection{Neural Networks as Arithmetic Circuits}\label{sec:nnasarith}
As mentioned before, \sys applies to neural networks that can be expressed as 
arithmetic circuits. This requirement places the following restrictions on 
the neural network layers.

\paragraph{Quadratic Activations}  
The activation functions in \sys  must be polynomials with integer coefficients (or, more precisely, coefficients 
in the field $\mathbb{F}_{p}$). The simplest of these is the element-wise quadratic 
activation function
whose output is simply the square of its input. 
Other commonly used activation functions such as ReLU, 
sigmoid or softmax activations
are precluded, \emph{except} in the final output 
layer. 
Prior work has shown that 
neural networks with quadratic 
activations have the same representation power as 
networks those with threshold activations and can be efficiently trained~\cite{gautier2016globally,livni2014computational}.

\paragraph{Sum Pooling}  
Pooling layers are commonly used to reduce the network size, to 
prevent overfitting and provide translation invariance. 
\sys uses
sum pooling, wherein the output of the pooling layer is the 
sum of activations
in each local region.  
%and average pooling, 
However, techniques such as max pooling~\cite{goodfellow2013maxout} 
and stochastic pooling~\cite{zeiler2013stochastic} are not supported 
since max and divisions operations are not easily represented as 
arithmetic circuits. 
%max operations are not efficiently supported in arithmetic circuits. 
%Further, since divisions 
%are not
%supported, stochastic pooling~\cite{} cannot be 
%used either, at least not directly.  
%average pooling instead 
%of sum
%pooling can be used during the training phase, but is effectively
%treated as sum pooling during inference by 

\paragraph{Finite Field Computations} \sys supports computations 
over elements of the field $\mathbb{F}_{p}$, that is, 
integers in the range $\{-\frac{p-1}{2}, \ldots, 0, \ldots, \frac{p-1}{2} \}$. The inputs, weights and 
all intermediate values computed in the network must 
lie in this range. 
Note that 
%divisions are not supported in arithmetic circuits, and 
due to the use 
of quadratic activations and sum pooling, the values in the network can become quite large. 
In practice, we will pick large primes to support these large values. 
We note that this restriction applies to the inference 
phase \emph{only}; the network can be trained with floating point inputs and weights. 
The inputs and weights are then re-scaled and quantized, 
as explained in Section~\ref{sec:implementation}, to finite field elements. 
%for instance $p=2_{61}-1$.

We note that the restrictions above are shared 
by a recently
proposed technique, CryptoNets~\cite{gilad2016cryptonets}, that seeks to 
perform neural network based inference on \emph{encrypted} 
inputs so as to guarantee data privacy. However, Cryptonets does not 
guarantee integrity and compared to \sys, 
incurs high costs for both the client and server (see Section~\ref{sec:runtimes} for a comparison). 
Conversely, \sys is targeted 
towards applications where integrity is critical, 
but does not provide privacy. 
%Our networks are a special class of polynomial neural networks that have been recently 
%analyzed by~\cite{gautier2016globally}.

\subsection{Mathematical Model}

An $L$ layer neural network with the constraints 
discussed above can be modeled, without loss of 
generality, as follows.
The input to the network is 
$\bm{x} \in \mathbb{F}_{p}^{n_{0} \times b}$, where 
$n_{0}$ is the dimension of each input and $b$ is the 
batch size. Layer $i \in [1,L]$ has $n_{i}$ output 
neurons\footnote{The $0^{th}$ layer is 
defined to be input layer and thus $\bm{y_{0}}=\bm{x}$.}, and is specified using a 
weight matrix $\bm{w}_{i-1} \in \mathbb{F}_{p}^{n_{i} \times n_{i-1}}$, and biases $\bm{b}_{i-1} \in  \mathbb{F}_{p}^{n_{i}}$.

The output of Layer $i \in [1,L]$, $\bm{y}_{i} \in  \mathbb{F}_{p}^{n_{i} \times b}$ is:
\begin{equation}\label{eq:layer}
 \bm{y}_{i} = \sigma_{quad}(\bm{w}_{i-1}.\bm{y}_{i-1} + \bm{b}_{i-1} \bm{1}^{T})
 \, \, \forall i \in [1,L-1]; \quad \bm{y}_{L} = \sigma_{out}(\bm{w}_{L-1}.\bm{y}_{L-1} + \bm{b}_{L-1}\bm{1}^{T}),
\end{equation}
where $\sigma_{quad}(.)$ is the quadratic activation function, $\sigma_{out}(.)$ is the activation function of the output layer, 
and $\bm{1} \in \mathbb{F}_{p}^{b}$ is the vector of all ones. 
We will typically use softmax activations in the output layer.  
We will also find it convenient to 
introduce the variable $\bm{z}_{i} \in \mathbb{F}_{p}^{{n_{i+1}} \times b} $ defined as
\begin{equation}\label{eq:layerz}
 \bm{z}_{i} = \bm{w}_{i}.\bm{y}_{i} + \bm{b}_{i}\bm{1}^{T}
 \, \, \forall i \in [0,L-1].
\end{equation}
The model captures both fully connected and convolutional layers; 
in the latter case the weight matrix is sparse. 
Further, without loss of generality,  
{all} successive 
linear
transformations in a layer, 
for instance sum pooling followed 
by convolutions, are represented using a 
single weight matrix. 

With this model in place, the goal of \sys is to enable 
the client to verify that  
$\bf{y_{L}}$ was correctly computed by the server.
We note that as in prior work~\cite{vu2013hybrid}, \sys amortizes the prover and 
verifier costs over 
\emph{batches} of inputs. If the server 
incorrectly computes the output corresponding to any input 
in a batch, the verifier rejects the entire batch of computations.

%% file: safetynets.tex
\section{\sys}
We now describe the design and implementation
of our end-to-end 
IP 
protocol for verifying execution of 
deep networks. 
The \sys protocol is a specialized
form of
the IP protocols developed by 
Thaler~\cite{thaler2013time} for verifying ``regular" 
arithmetic circuits, that themselves specialize and refine
prior work~\cite{cmt}. The starting point for the protocol is a polynomial 
representation of the network's inputs and parameters, referred to as a multilinear extension. 

\paragraph{Multilinear Extensions} 
Consider a matrix 
$\bm{w} \in \mathbb{F}_{p}^{n \times n}$. 
Each row and column of $\bm{w}$ can be referenced using 
$m=\log_{2}(n)$ bits, and consequently one can 
represent $\bm{w}$ as a function $W: \{0,1\}^{m} \times \{0,1\}^{m} \rightarrow \mathbb{F}_{p}$. 
That is, given Boolean vectors 
$\bm{t}, \bm{u} \in \{0,1\}^{m}$, the function
$W(\bm{t}, \bm{u})$ returns the element of $\bm{w}$ 
at the row and column specified by Boolean vectors 
$\bm{t}$ and  $\bm{u}$, respectively. 

A \emph{multi-linear extension} of $W$ is a polynomial 
function $\tilde{W}: \mathbb{F}_{p}^{m} \times \mathbb{F}_{p}^{m} \rightarrow \mathbb{F}_{p}$ that has the following two properties:
(1) given vectors 
$\bm{t}, \bm{u} \in \mathbb{F}_{p}^{m}$
such that $\tilde{W}(\bm{t}, \bm{u})  = W(\bm{t}, \bm{u})$ for all 
points on the unit hyper-cube, that is, for all 
$\bm{t}, \bm{u} \in \{0,1\}^{m}$; and (2) $\tilde{W}$ has 
degree 1 in each of its variables. 
In the remainder of this discussion, we will 
use $\tilde{X}$, $\tilde{Y_{i}}$ and
$\tilde{Z_{i}}$ and
$\tilde{W_{i}}$ to refer to multi-linear extensions of 
$\bm{x}$, $\bm{y_{i}}$, $\bm{z_{i}}$, and $\bm{w_{i}}$, respectively, for $i \in [1,L]$. 
We will also assume, for clarity of exposition, that 
the biases, $\bm{b_{i}}$ are 
zero for all layers. 
The supplementary draft
describes how biases are incorporated. 
Consistent with the IP literature, the description of 
our protocol refers to the client as the verifier and 
the server as the prover. 

\paragraph{Protocol Overview}
The verifier seeks to check the result $\bm{y}_{L}$ 
provided by the prover 
corresponding to input $\bm{x}$. 
Note that $\bm{y}_{L}$ is the output of the final activation 
layer which, as discussed in Section~\ref{sec:nnasarith}, is the only 
layer that does not use quadratic 
activations, and is hence not amenable to an IP. 

Instead, in \sys, the prover computes and 
sends  
$\bm{z}_{L-1}$ (the \emph{input} of the final activation 
layer) as a result to the verifier. $\bm{z}_{L-1}$  has the 
same dimensions as $\bm{y}_{L}$ and therefore this 
refinement has 
no impact on the server to client bandwidth. Furthermore, the verifier can 
easily compute $\bm{y}_{L} = \sigma_{out}(\bm{z}_{L-1})$ locally.

Now, the verifier needs to check whether the prover
computed $\bm{z}_{L-1}$ correctly. As noted by 
Vu et al.~\cite{vu2013hybrid}, this check can be replaced by a check 
on whether the multilinear extension of $\bm{z}_{L-1}$
is correctly computed at a randomly picked point in the field, 
with minimal impact on the soundness error. 
That is, the verifier 
picks two vectors, $\bm{q}_{L-1} \in \mathbb{F}_{p}^{log(n_{L})}$ and 
$\bm{r}_{L-1} \in \mathbb{F}_{p}^{log(b)}$ 
at \emph{random}, 
evaluates 
$\tilde{Z}_{L-1}(\bm{q}_{L-1},\bm{r}_{L-1})$, and checks 
whether it was correctly computed using a 
specialized sum-check protocol for matrix multiplication 
due to Thaler~\cite{thaler2013time}
(described in Section~\ref{sec:sumcheckmm}).
%Equation~\ref{eq:}).  

Since $\bm{z}_{L-1}$ depends on $\bm{w}_{L-1}$ and 
$\bm{y}_{L-1}$, sum-check yields assertions on the values of $\tilde{W}_{L-1}(\bm{q}_{L-1},\bm{s}_{L-1})$ and
$\tilde{Y}_{L-1}(\bm{s}_{L-1},\bm{r}_{L-1})$, where 
$\bm{s}_{L-1} \in \mathbb{F}_{p}^{log(n_{L-1})}$ is 
another random vector picked by the verifier during sum-check. 

$\tilde{W}_{L-1}(\bm{q}_{L-1},\bm{s}_{L-1})$ is an 
assertion about the weight of the final layer. 
This is checked 
by the verifier locally since the weights are known to both 
the prover and verifier. 
Finally, the verifier uses 
our specialized sum-check protocol for activation 
layers (described in Section~\ref{sec:sumcheckquad}) 
to reduce 
the assertion on
$\tilde{Y}_{L-1}(\bm{s}_{L-1},\bm{r}_{L-1})$ to an 
assertion on $\tilde{Z}_{L-2}(\bm{q}_{L-2},\bm{s}_{L-2})$.
The protocol repeats till 
it reaches the input layer and produces an 
assertion on 
$\tilde{X}(\bm{s}_{0},\bm{r}_{0})$, the multilinear 
extension of the input $\bf{x}$. The verifier checks this 
locally. If at any point in the protocol, the verifier's 
checks fail, it rejects the prover's computation. 
Next, we describe the
sum-check protocols for matrix multiplication and 
activation that \sys uses.

%The goal of the \sys protocol is to enable the 
%verifier check the 
%correctness of the output $\bm{y_{L}}$ 
%$\bm{x}$, 

\subsection{Sum-check for Matrix Multiplication}\label{sec:sumcheckmm}
Since $\bm{z}_{i} = \bm{w}_{i}.\bm{y}_{i}$ 
(recall we assumed zero biases for clarity), 
we can check an assertion about the multilinear extension 
of
$\bm{z}_{i}$ evaluated at randomly picked 
points $\bm{q}_{i}$ and 
$\bm{r}_{i}$ by expressing $\tilde{Z}_{i}(\bm{q}_{i},\bm{r}_{i})$ 
as~\cite{thaler2013time}:
\begin{equation}\label{ref:z}
\tilde{Z}_{i}(\bm{q}_{i},\bm{r}_{i}) = \sum_{j\in\{0,1\}^{\log(n_i)}} \tilde{W}_{i}(\bm{q}_{i},\bm{j}).
\tilde{Y}_{i}(\bm{j},\bm{r}_{i})
\end{equation}
Note that Equation~\ref{ref:z} has the same form as 
the sum-check problem in Equation~\ref{eq:claim1}. Consequently 
the sum-check protocol described in Section~\ref{sec:ip}
can be used to verify this 
assertion. At the end of the sum-check rounds, 
the verifier will have assertions
on $\tilde{W}_{i}$ which it checks locally,  
and $\tilde{Y}_{i}$ 
which is checked using the sum-check protocol for 
quadratic activations described in Section~\ref{sec:sumcheckquad}. 

The prover run-time for
running the sum-check protocol in  
layer $i$ is $\mathbb{O}(n_i (n_{i-1}+b))$, 
the verifier's run-time is $\mathbb{O}(n_i n_{i-1})$
and the prover/verifier 
exchange $4\log(n_i)$ field elements.

\subsection{Sum-check for Quadratic Activation}\label{sec:sumcheckquad}
In this step, we check an assertion about the 
output of quadratic activation layer $i$, $\tilde{Y}_{i}(\bm{s}_{i},\bm{r}_{i})$, by writing 
it in terms of the input of the activation layer as follows:
\begin{equation}\label{ref:act}
\tilde{Y}_{i}(\bm{s}_{i},\bm{r}_{i}) = \sum_{j\in\{0,1\}^{\log(n_i)}, k\in\{0,1\}^{\log(b)} } 
\tilde{I}(\bm{s}_{i},\bm{j})
\tilde{I}(\bm{r}_{i},\bm{k})
\tilde{Z}_{i-1}^{2}(\bm{j},\bm{k}),
\end{equation}
where $\tilde{I}(.,.)$ is the multilinear extension of the 
identity matrix. Equation~\ref{ref:act} 
can also be verified using the sum-check protocol, and 
yields an assertion about
$\tilde{Z}_{i-1}$, i.e., the inputs to the activation layer. 
This assertion is in turn checked using the protocol described in Section~\ref{sec:sumcheckmm}. 

The prover run-time for
running the sum-check protocol in  
layer $i$ is $\mathbb{O}(bn_i)$, 
the verifier's run-time is $\mathbb{O}(\log(bn_{i}))$
and the prover/verifier 
exchange $5\log(bn_i)$ field elements.
This completes the theoertical description of the \sys specialized
IP protocol.

\begin{lemma}
The \sys verifier rejects incorrect computations with 
probability
greater than 
%$\left( 1 - \frac{d}{p} \right) _{n} \approx 
$\left( 1 - \epsilon\right)$ 
where $\epsilon = \frac{3b\sum_{i=0}^{L}n_{i}}{p}$ is the soundness error. 
%\frac{2\sum_{i=0}^{L-1}n_{i} + 3\sum_{i=1}^{L-1}bn_{i} + bn_{L}}{p}$. 
% for $p>>d$. 
\end{lemma}
In practice, with $p=2^{61}-1$ the soundness error $<\frac{1}{2^{30}}$ for practical 
network parameters and batch sizes.

\subsection{Implementation}\label{sec:implementation}
The fact that \sys operates only 
on elements in a finite field
$\mathbb{F}_{p}$ 
during inference imposes a practical challenge. 
That is, how do we convert floating point
inputs and weights from training 
into field elements, and how do we select the size of the 
field $p$? 

Let 
$\bm{w'}_{i} \in  \mathbb{R}^{n_{i-1} \times n_{i}}$
and $\bm{b'}_{i} \in  \mathbb{R}^{n_{i}}$ be the floating 
point parameters obtained from training for each layer  $i \in [1,L]$. We convert the weights to integers by multiplying 
with a constant 
$\beta > 1$ and rounding, i.e., 
$\bm{w}_{i} = \lfloor  \beta \bm{w'}_{i} \rceil$. 
We do the same for inputs with a scaling factor 
$\alpha$, i.e., 
$\bm{x} = \lfloor  \alpha \bm{x'} \rceil$.
Then, to ensure that all values in the network scale 
isotropically, we must set $
\bm{b}_{i} = \lfloor \alpha^{2^{i-1}} \beta^{(2^{i-1}+1)} \bm{b'}_{i} \rceil$. 

While larger 
$\alpha$ and 
$\beta$ values imply lower quantization errors, 
they also result in large values in the network, especially 
in the layers closer to the output. Similar 
empirical 
observations were made by the CryptoNets work~\cite{gilad2016cryptonets}.
To ensure accuracy the values in the network 
must lie 
in the range $[-\frac{p-1}{2},\frac{p-1}{2}]$; this influences the 
choice of the prime $p$. 
On the other hand, we note that large primes increase the 
verifier and prover run-time because of the higher 
cost of performing modular additions and 
multiplications. 

As in prior works~\cite{cmt, thaler2013time,vu2013hybrid}, we restrict our choice of $p$ to Mersenne primes since they afford efficient modular arithmetic implementations, and 
specifically to the primes $p=2^{61}-1$ and 
$p=2^{127}-1$. 
For a given $p$, we explore and 
different
values of $\alpha$ and $\beta$ and use the 
validation dataset to the pick the ones
that maximize accuracy 
while ensuring that the values in the network 
lie within $[-\frac{p-1}{2},\frac{p-1}{2}]$.

\begin{comment}
Noting that 
We refine the CMT protocol to apply to the square activation layer. The multilinear extension of the output of this layer can be written as 
$$\tilde{Z}(p_1)=\sum_{p_2\in\{0,1\}^{\log n}} \tilde{I}(p_1,p_2)\tilde{Y}_{2}(p_2)$$
where $p_1\in\mathbb{F}_{\log n}$ and $\tilde{Y}$ is the multilinear extension of the input to the activation layer. We define $\tilde{I}(a,b) = \prod_{i=1}_{\log n}((1-a_i).(1-b_i)+a_ib_i)$.
To verify this layer, the sum-check protocol is applied to the polynomial $g_{p_1}(p_2)=\tilde{I}(p_1,p_2)\tilde{Y}_{2}(p_2)$.
\end{comment}

%% file: eval.tex
\section{Empirical Evaluation} \label{eval}

In this section, 
%we benchmark the 
%run-time of 
%\sys verifier and prover and 
we present
empirical evidence to support our 
claim that \sys enables low-cost
verifiable execution of deep neural networks on untrusted 
clouds without compromising classification accuracy.
%practicality of \sys. Since \sys 

\subsection{Setup}

\paragraph{Datasets} We evaluated \sys on three classifications tasks. 
(1) Handwritten digit recognition on the MNIST  dataset, using 50,000 training, 10,000 validation and 
10,000 test images. (2) A more challenging version of digit recognition, MNIST-Back-Rand, 
an artificial dataset generated by inserting a random background into MNIST image~\cite{mnistvariation}. 
The dataset has 10,000 training, 2,000 validation and 50,000 test images. 
ZCA whitening is applied to the raw dataset before training and testing~\cite{coates2011analysis}. 
(3) Speech recognition on the TIMIT dataset, split into a training 
set with 462 speakers, a validation set with 144 speakers and a testing set with 24 speakers. 
The raw audio samples are pre-processed as described by~\cite{ba2014deep}. Each example includes 
its preceding and succeeding 7 frames, resulting in a 1845-dimensional input in total.
During testing, all labels are mapped to 39 classes~\cite{lee1989speaker} for evaluation.
%MNIST and MNIST-back-rand, 

\paragraph{Neural Networks} For the two MNIST tasks, we used a convolutional neural network same as~\cite{zhang2016convexified}
with 2 convolutional layers with $5\times5$ filters, a stride of $1$ and a 
mapcount of 16 and 32 for the first and second layer respectively. 
Each convolutional layer is followed by quadratic activations and $2\times2$ 
sum pooling with a stride of 2. The fully connected layer uses softmax activation. We refer to this network as 
\textbf{CNN-2-Quad}. 
For TIMIT, we use a four layer network described by ~\cite{ba2014deep} with 3 hidden,  
fully connected layers with $2000$ neurons and quadratic activations.
The output layer is fully connected with $183$ output neurons and softmax activation.
We refer to this network as \textbf{FcNN-3-Quad}.
Since quadratic activations are not commonly used, we compare the performance of CNN-2-Quad and 
FcNN-3-Quad with baseline versions in which the quadratic activations are replaced by ReLUs. The baseline 
networks are \textbf{CNN-2-ReLU} and \textbf{FcNN-3-ReLU}.

%TIMIT is a speechN-2-Quad with a network in which the q recognition datauadrarset. It is . The raw audio data are broke into frames using 25ms hamming window shifting by 10ms. From each frame, 40 MFCCs are extracted, along with energy and their first and second derivatives\cite{ba2014deep}. With nearby +/- 7 frames included, each example has 1845 dimensions in total. It finally gives us 1.1M examples in training set, 330k in validation set and 50k in testing set. We use three states for each of the 61 phoneme labels during training. Thus we get 183 target classes. 

The hyper-parameters for training are selected based on the validation datasets.
The Adam Optimizer is used for CNNs with learning rate 0.001, exponential decay and dropout probability 0.75. 
The AdaGrad optimizer is used for FcNNs with a learning rate of 0.01 and dropout probability 0.5.
We found that norm gradient clipping 
was required for training the CNN-2-Quad and FcNN-3-Quad networks, since the gradient values for quadratic 
activations can become large.

%In training the CNN, dropout with probability of 0.75 is added after each convolutional layer. We choose Adam as the optimizer and update the learning rate using the following formula: 
%$$
%lr = lr_{min} + (lr_{max} - lr_{min})*e^{\frac{t}{decay}}
%$$
%where $lr_{min}$ is set to 1e-3, $lr_{max}$ is set to 3e-2, $decay$ is set to 2000 and $t$ is the iteration. In training the FCNN, dropout with probability of 0.5 with is added after each hidden layer. We choose AdaGrad as the optimizer and the initial learning rate was set to 1e-2. No learning rate decay is used. For both FCNN and CNN, gradient clipping is applied when using quadratic activation function. The norm of the gradients is clipped to 100 in CNN and 1 in FCNN.

Our implementation of \sys uses Thaler's code for the IP protocol for
matrix multiplication ~\cite{thaler2013time} and our own implementation of the IP for quadratic 
activations. We use an Intel Core i7-4600U CPU running at $2.10$ GHz for benchmarking.

\subsection{Classification Accuracy of \sys} 

\begin{figure}
\centering
\subfloat[MNIST]{\includegraphics[width=0.35\textwidth]{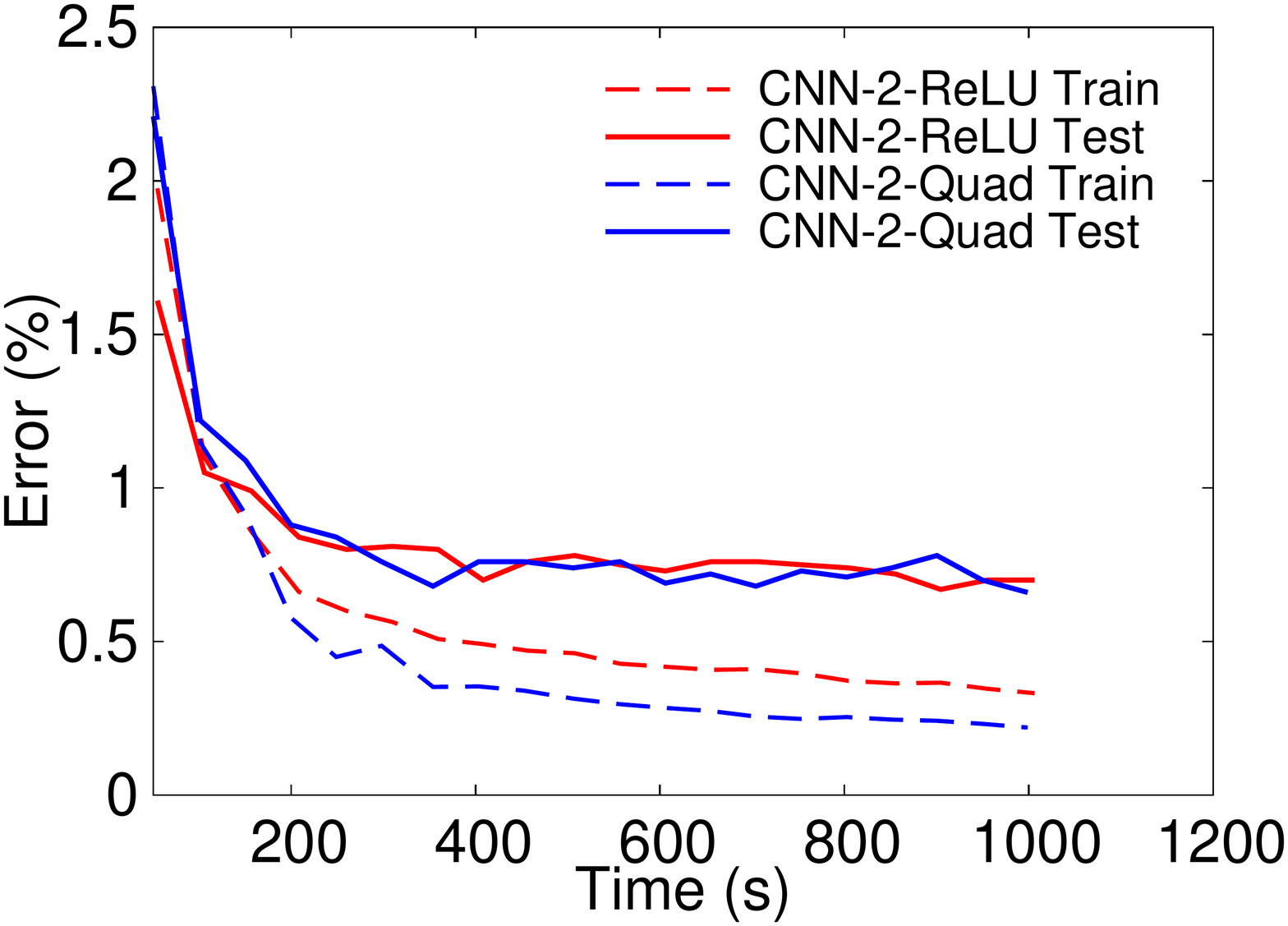}\label{fig:mnistbasic}}
\subfloat[MNIST-Back-Rand]{\includegraphics[width=0.35\textwidth]{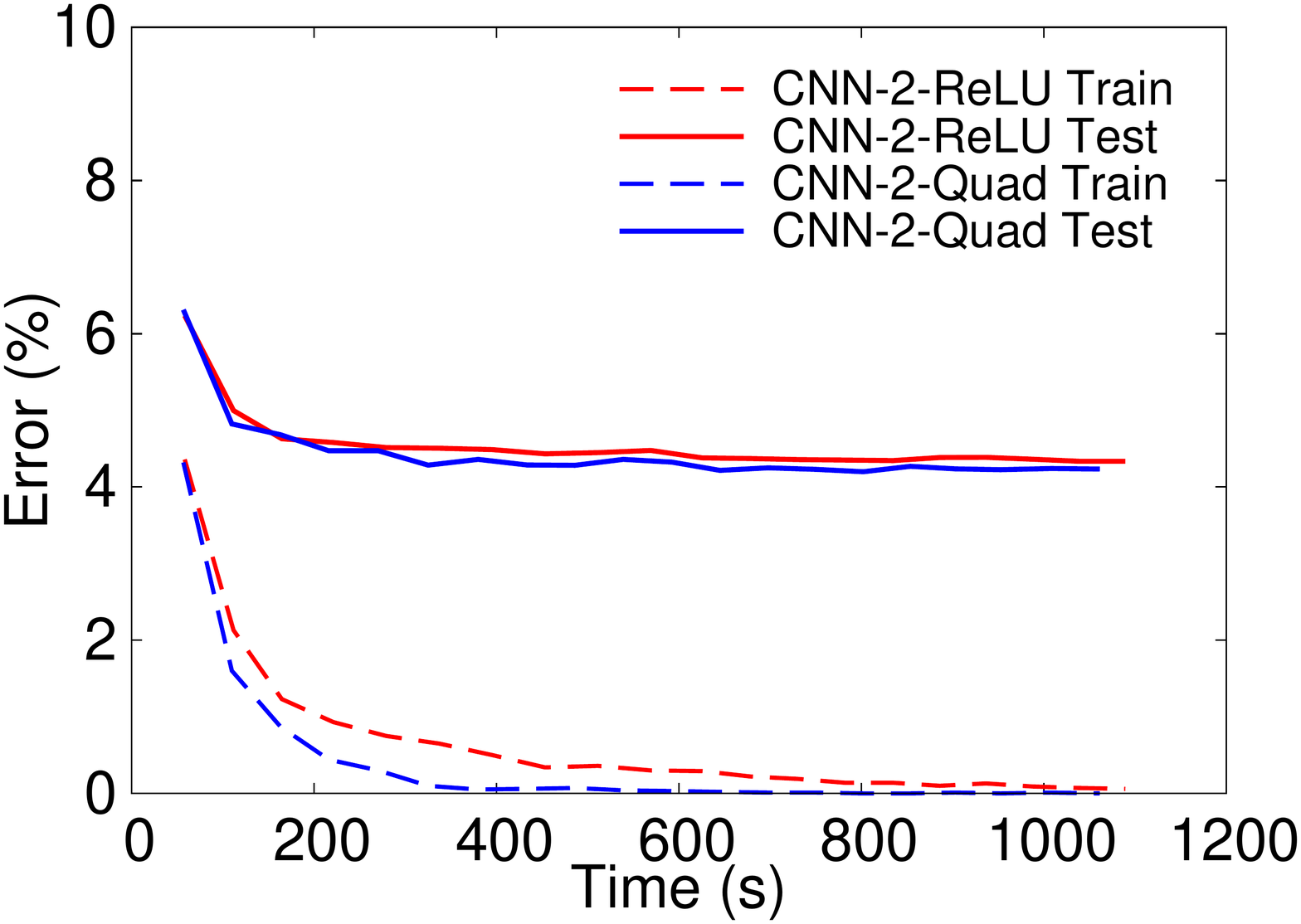}\label{fig:mnistrand}}
\subfloat[TIMIT]{\includegraphics[width=0.35\textwidth]{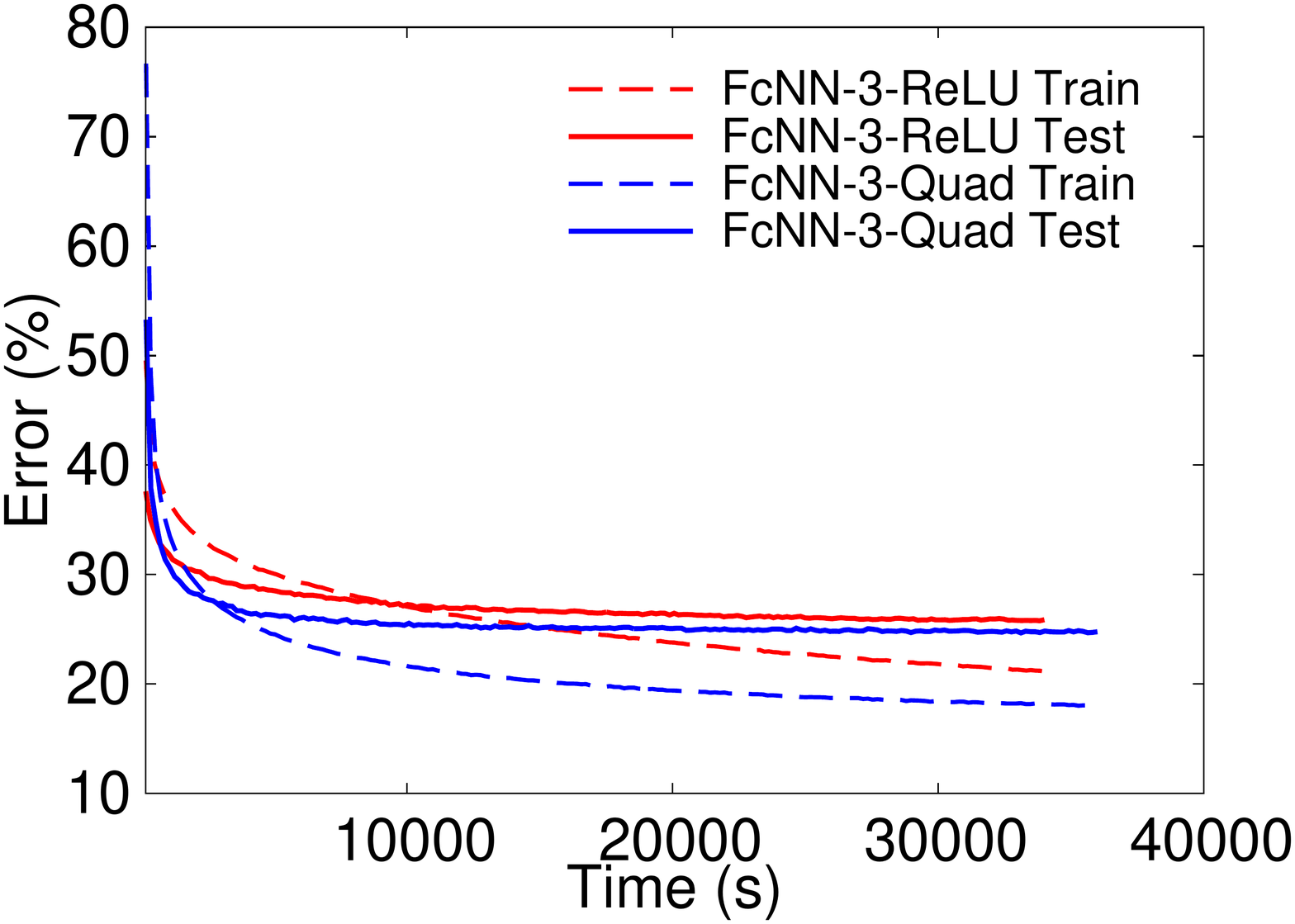}\label{fig:timit}}
\caption{Evolution of training and test error for the MNIST, MNIST-Back-Rand and TIMIT tasks.}
\label{fig:reluquad}
\vspace{-0.15in}
\end{figure}

\sys places certain restrictions on the activation function
(quadratic) and requires weights and inputs to be integers (in field $F_{p}$). 
We begin by analyzing how (and if) 
these restrictions impact classification accuracy/error. 
Figure~\ref{fig:reluquad} compares training and test error of 
CNN-2-Quad/FcNN-3-Quad
versus CNN-2-ReLU/FcNN-3-ReLU. For all three tasks, the networks with quadratic 
activations are competitive with 
networks that use ReLU activations.
Further, we observe that the 
networks with quadratic activations 
appear to converge faster during training, 
possibly because their gradients are larger despite gradient clipping. 
%on MNIST and MNIST-Back-Rand, 
%and  with on TIMIT. 

Next, we used the scaling and rounding
strategy proposed in Section~\ref{sec:implementation} to convert
weights and inputs to integers.  
Table~\ref{tab:quant} shows the impact of scaling factors 
$\alpha$ and 
$\beta$ 
on the classification error and maximum values observed in the network during inference for
MNIST-Back-Rand. 
The validation 
error drops as $\alpha$
and $\beta$ are increased. 
On the other hand, for $p=2^{61}-1$, 
the largest value allowed is 
$1.35\times10^{18}$; this rules out $\alpha$ and $\beta$  greater 
than $64$, as shown in the table. 
For MNIST-Back-Rand, we pick 
$\alpha=\beta=16$ based on validation data, and obtain a test error 
of $4.67\%$.
Following a similar methodology, we obtain a test error of 
$0.63\%$  for MNIST 
($p=2^{61}-1$) and $25.7\%$  for 
TIMIT ($p=2^{127}-1$). We note that 
\sys does not support  
techniques 
such as Maxout~\cite{goodfellow2013maxout} that
have demonstrated 
lower error on MNIST ($0.45\%$).  
Ba et al.~\cite{ba2014deep} report an error of $18.5\%$ for TIMIT using an 
\emph{ensemble} of nine deep neural networks, which \sys might be able to support
by verifying each network individually and performing ensemble averaging at the client-side. 
%\todo{Tianyu:whats state of the art?}, 
%but in return, 
%achieves low-cost verifiabilty of outsourced execution. 
\vspace{-7pt}
\subsection{Verifier and Prover Run-times}\label{sec:runtimes}
\vspace{-7pt}

\begin{wrapfigure}{l}{0.4\textwidth}
    %\subfloat[CNN-2-Quad]{
    \centering
        \includegraphics[width=0.4\textwidth]{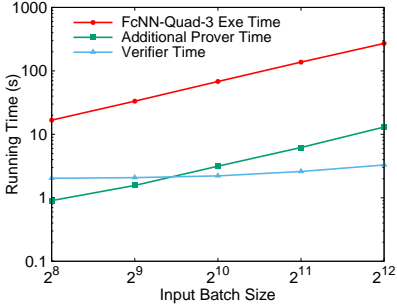}
    %}
    %\subfloat[FcNN-3-Quad]{
    %    \includegraphics[width=0.3\textwidth]{figures/timit_ver}
    %}
    \caption{Run-time of verifier, prover and baseline execution time for the 
    arithmetic circuit representation of FcNN-Quad-3 versus input batch size.}
    \label{fig:mnist_ver}
\end{wrapfigure}

The relevant performance metrics for 
\sys are (1) the client's (or verifier's) 
run-time, (2) the server's run-time which includes baseline time to 
execute the neural network and overhead of generating proofs, and (3) the bandwidth required 
by the IP protocol. Ideally, these quantities should be small, and importantly, the 
client's  run-time
should be smaller than the case in which  
it executes the network by itself. 
Figure~\ref{fig:mnist_ver} plots run-time data over  
input batch sizes ranging from 256 to 2048 
for FcNN-Quad-3.

For FcNN-Quad-3, the client's time for verifying proofs 
is $8\times$ to $82\times$ faster than the baseline in which it 
executes FcNN-Quad-3 itself, and decreases with batch size. 
The increase in the server's execution time due to the overhead of 
generating proofs is only
$5\%$ over the baseline unverified execution of FcNN-Quad-3.
The prover and verifier exchange less than 8 KBytes of data during the  
IP protocol for a batch 
size of 2048, which is  
negligible (less than $2\%$)
compared to the bandwidth 
required to communicate inputs 
and outputs back and forth. 
In all settings, the soundness error $\epsilon$, i.e., 
the chance that the verifier fails to detect incorrect computations by the server
is less than $\frac{1}{2^{30}}$, a negligible value. 
%The results for CNN-Quad-3 
%are qualitatively similar. 
%One point of comparison for \sys is a
%solution 
We note \sys has significantly 
lower bandwidth costs compared to an approach that \emph{separately} verifies
the execution of each layer
using only the IP protocol for matrix multiplication. 
%and requires 
%the server to send the inputs/outputs of each activation layer to the client.

A recent and closely related technique, CryptoNets~\cite{gilad2016cryptonets}, 
uses homomorphic 
encryption to provide privacy, but not integrity, 
for neural networks executing in the cloud. 
Since \sys and CryptoNets target different security goals
%, integrity and privacy, 
%respectively, 
a direct comparison is not entirely meaningful. However, from the data 
presented in the CryptoNets paper, 
we note that the client's run-time for MNIST using a CNN similar to ours and an input batch size $b=4096$ is about 600 seconds, 
primarily because of the high cost of encryptions.
For the same batch size, the client-side run-time of \sys is less than 
10 seconds. From this rough comparison, it appears that integrity is a more practically achievable security goal than privacy.

\begin{table}[]
\centering
\caption{Validation error and maximum value observed in the network for MNIST-Rand-Back 
and different values of scaling parameters, $\alpha$ and $\beta$. Shown in bold red font are values of 
$\alpha$ and $\beta$ that are infeasible because the maximum value exceeds that allowed by prime 
$p=2^{61}-1$.}
\label{tab:quant}
\resizebox{\textwidth}{!}{
\begin{tabular}{|l|l|l|l|l|l|l|l|l|l|l|}
\hline
\multicolumn{1}{|c|}{}                    & \multicolumn{2}{c|}{$\alpha=4$} & \multicolumn{2}{c|}{$\alpha=8$} & \multicolumn{2}{c|}{$\alpha=16$} & \multicolumn{2}{c|}{$\alpha=32$}                         & \multicolumn{2}{c|}{$\alpha=64$}                         \\ \cline{2-11} 
\multicolumn{1}{|c|}{\multirow{-2}{*}{$\beta$}} & Err       & Max       & Err       & Max        & Err       & Max         & Err   & Max                                     & Err   & Max                                     \\ \hline
4                                         & $0.188$     & $4.0\times 10^8$    & $0.073$     & $4.0\times 10^{10}$    & $0.042$     & $5.5\times 10^{12}$     & $0.039$ & $6.6\times 10^{14}$                                 & $0.04$  & $8.8\times 10^{16}$                                 \\ \hline
8                                         & $0.194$     & $6.1\times 10^9$    & $0.072$     & $6.9\times 10^{11}$    & $0.039$     & $8.3\times 10^{13}$     & $0.038$ & $1.0\times 10^{16}$                                 & $0.037$ & {\color[HTML]{FE0000} $\bm{1.3\times 10^{18}}$} \\ \hline
16                                        & $0.188$     & $9.4\times 10^{10}$    & $0.072$     & $1.1\times 10^{13}$    & $0.036$     & $1.3\times 10^{15}$     & $0.037$ & $1.6\times 10^{17}$                                 & $0.035$ & {\color[HTML]{FE0000} $\bm{2.1\times 10^{19}}$} \\ \hline
32                                        & $0.186$     & $1.5\times 10^{12}$    & $0.073$     & $1.7\times 10^{14}$    & $0.038$     & $2.1\times 10^{16}$     & $0.037$ & {\color[HTML]{FE0000} $\bm{2.6\times 10^{18}}$} & $0.036$ & {\color[HTML]{FE0000} $\bm{3.5\times 10^{20}}$} \\ \hline
64                                        & $0.185$     & $2.5\times 10^{13}$    & $0.073$     & $2.8\times 10^{15}$    & $0.038$     & $3.4\times 10^{17}$     & $0.037$ & {\color[HTML]{FE0000} $\bm{4.2\times 10^{19}}$} & $0.036$ & {\color[HTML]{FE0000} $\bm{5.6\times 10^{21}}$} \\  \hline
\end{tabular}}
\end{table}

\begin{comment}
a input quantization factor
b weight quantization factor
mnist original
a=8 b=16
valid error = 0.78 (non quantized = 0.76)
test error = 0.63
nonquantized test error = 0.66

mnist-rand
a=16 b=16
valid error = 3.6 (non quantized = 3.56)
test error=4.67
nonquantized test error = 4.23

timit
a=4 b=64
valid error = 25.76 (non quantized = 24.81)
test error=25.77
non quantized test error = 24.8
\end{comment}

\vspace{-10pt}

\section{Conclusion}
\vspace{-7pt}
In this paper, we have presented \sys, a new framework that 
allows a client to provably verify the correctness of deep neural network 
based inference running on an untrusted clouds. Building upon the rich literature on 
interactive proof systems for verifying general-purpose and specialized computations, 
we designed and implemented a specialized IP protocol tailored for 
a certain class of deep neural networks, i.e., those that can be represented as arithmetic 
circuits. We showed that placing these restrictions did not impact the accuracy of the 
networks on real-world classification
tasks like digit and speech recognition, while enabling a client to verifiably 
outsource 
inference to the cloud at low-cost. For our future work, we will apply \sys to deeper networks and extend it to address \emph{both} integrity and privacy. There
are VC techniques~\cite{parno2013pinocchio} that guarantee both, but typically come at higher costs.

\vspace{-7pt}

\begin{comment}
Our future work will address some of the limitations of \sys. 
First, \sys has trouble scaling to very deep networks, because the values in the network grow larger and 
necessitate the use of even larger primes.  
Further, \sys does not apply to 
setting in which \emph{both} integrity and privacy are desired. Other VC techniques that
do provide both integrity and privacy guarantees~\cite{parno2013pinocchio} are worth exploring, but typically 
come at higher costs. 
\end{comment}

%% file: proofs.tex
\section*{Proof of Lemma 3.1}
\paragraph{Lemma 3.1}
The \sys verifier rejects incorrect computations with 
probability
greater than 
%$\left( 1 - \frac{d}{p} \right) _{n} \approx 
$\left( 1 - \epsilon\right)$ 
where $\epsilon = \frac{3b\sum_{i=0}^{L}n_{i}}{p}$ is the soundness error. 
%\frac{2\sum_{i=0}^{L-1}n_{i} + 3\sum_{i=1}^{L-1}bn_{i} + bn_{L}}{p}$. 
% for $p>>d$. 

\begin{proof}
Verifying a multi-linear extension of the output sampled at a random 
point, instead of each value adds a soundness error of $\epsilon=\frac{bn_{L}}{p}$.
Each instance of the sum-check protocol adds to the soundness error~\cite{vu2013hybrid}. 
The IP protocol for 
matrix multiplication adds a soundness error of $\epsilon=\frac{2n_{i-1}}{p}$ in Layer $i$~\cite{thaler2013time}.
Finally, the IP protocol for quadratic activations adds a soundness error of $\epsilon=\frac{3bn_{i}}{p}$ in 
Layer $i$~\cite{thaler2013time}.
Summing together we get a total soundness error 
of $\frac{2\sum_{i=0}^{L-1}n_{i} + 3\sum_{i=1}^{L-1}bn_{i} + bn_{L}}{p}$.
The final result is an upper bound on this value.
\end{proof}

\section*{Handling Bias Variables}
We assumed that the bias variables were zero, allowing us to write 
$bm{z}_{i} = \bm{w}_{i}.\bm{y}_{i}$ while it should be 
$bm{z}_{i} = \bm{w}_{i}.\bm{y}_{i} + \bm{b}_{i}\bm{1}^{T}$. 
Let  $\bm{z'}_{i} = \bm{w}_{i}.\bm{y}_{i}$ We seek to convert an assertion 
on $\tilde{Z}_{i}(\bm{q}_{i},\bm{r}_{i})$ to an assertion on
$\tilde{Z'}_{i}$. We can do so by noting that:
\begin{equation}
\tilde{Z}_{i}(\bm{q}_{i},\bm{r}_{i}) = \sum_{j\in\{0,1\}^{\log(n_i)}} \tilde{I}(\bm{j},\bm{q}_{i}) ( \tilde{Z'}_{i}(\bm{j},\bm{r}_{i}) + 
\tilde{B}_{i}(\bm{j}) )
\end{equation}
which can be reduced to sum-check and thus yields an assertion on  $\tilde{B}_{i}$ 
which the verifier checks locally and 
$\tilde{Z'}_{i}$, which 
is passed to the IP protocol for matrix multiplication.